\documentclass[11pt, a4paper]{scrartcl}
\usepackage[T1]{fontenc}
\usepackage{latexsym}
\usepackage{amssymb}
\usepackage[nosumlimits]{amsmath}
\usepackage{graphicx}
\usepackage{natbib}
\usepackage{float}
\usepackage{url}

\usepackage{xcolor}
\usepackage[pdftex,
              pdftitle={Recalibrating Binary Probabilistic Classifiers},
               pdfsubject={Machine learning, credit risk management},
               pdfkeywords={},
               pdfauthor={Dirk Tasche},
               pdfstartview=FitR,
               breaklinks=true,
               colorlinks=true,
               citecolor=teal,
               linkcolor=magenta,
               urlcolor=olive]{hyperref}
               
\newtheorem{theorem}{Theorem}[section]

\newtheorem{remark}[theorem]{Remark}

\newtheorem{proposition}[theorem]{Proposition}

\begin{document}
\title{Recalibrating Binary Probabilistic Classifiers}
\author{Dirk Tasche\\
Centre for Business Mathematics and Informatics\\ North-West University, South
Africa}

\date{}
\maketitle              % typeset the header of the contribution
\begin{abstract}
Recalibration of binary probabilistic classifiers to a target prior probability is an important task in areas like
credit risk management. However, recalibration of a classifier learned on a training dataset to a target  
on a test dataset in general is not a 
well-defined problem because there might be more than one way to transform the original
posterior probabilities such that the target is matched. In this paper, methods for recalibration are analysed from a distribution shift
perspective. Distribution shift assumptions linked to the 
area under the curve (AUC) of a probabilistic classifier are found to be useful for
the design of meaningful recalibration methods. Two new methods called parametric covariate shift
with posterior drift (CSPD) and ROC-based quasi moment matching (QMM) are proposed and tested 
together with some other methods in an example setting. The outcomes of the test suggest that the QMM methods
discussed in the paper can provide appropriately conservative results in evaluations with concave functions
like for instance risk weights functions for credit risk. 


\noindent\textbf{MSC classes:} 68T09, 91G40
\end{abstract}

%%%%%%
\section{Introduction}
%%%%%%

Occasionally binary probabilistic classifiers are learned on a training dataset 
(typically a collection of data points observed in the past) and then are
applied to a test dataset which reflects a joint distribution of features and labels different than
the distribution of the training dataset. Actually, quite often it is unknown how much the
training and test distributions differ because for the instances in the test dataset only 
the features but not the labels can be observed. Hence, the feature training and test distributions might
be different while the posterior probabilities are identical for the training and test datasets. 
This kind of dataset shift is called covariate shift and is rather benign in 
principle as it would not require any change of the
probabilistic classifier \citep{storkey2009training}. 
If however another type of dataset shift other than covariate shift is incurred, applying the
classifier learned on the training dataset to the instances in the test dataset without any
change risks to generate unreliable predictions of the labels.

This paper aims to study the situation where indeed no labels are observed in the test dataset but
where there exists an estimate of the proportion of positive labels, i.e.\ an estimate of the
test prior probability of the positive class. The problem is then to \emph{recalibrate}
the probabilistic classifier learned on the training dataset such that the mean of the
recalibrated classifier on the test dataset matches the estimate of the proportion of positive labels.

This is a common situation in credit risk management, 
see for instance Chapter~4 of \citet{Bohn&Stein}.
So called probabilities of default (PDs) are estimated on a training sample with
observations of solvent and defaulted borrowers and must be recalibrated before being evaluated
for the borrowers in a live portfolio. The future solvency states of these borrowers in general
are unknown but typically an estimate of the proportion of borrowers who are going to default
is available. Sometimes, such estimates are conservative, i.e.~they are likely to significantly overestimate
the proportion of defaulters. Conservatism of the estimates can be a regulatory requirement or it
could be part of stress testing exercises intended to assess the impact of unfavourable economic
circumstances on the portfolio.

Recalibration of posterior probabilities learned on a training dataset to a target prior probability 
on a test dataset is not a 
well-defined problem because there is more than one way to transform the original
posterior probabilities such that the target is matched. Therefore, as an additional criterion
to identify meaningful solutions the impact of the recalibration
method on the values of concave or nearly concave functions 
of the posterior probabilities is investigated. The risk weight
functions for the calculation of minimum required capital under the ``internal ratings based (IRB) approach'' of the
Basel credit risk framework \citep{BCBSFramework}
are a primary example of such functions. The findings of this paper
can inform the choice of the recalibration method in credit risk management and similar contexts.

This paper is organised as follows:
Section~\ref{se:relWork} puts the paper into the context of related work.
Section~\ref{se:setting} describes the technical details of the setting of the paper and the recalibration problem.
In addition, it introduces the evaluation of the solutions with a concave function as a criterion for 
assessing their appropriateness.
Section~\ref{se:approaches} presents a number of methods for recalibration. Parametric CSPD (Section~\ref{se:CSPD})
and ROC-based QMM (Section~\ref{se:robLogit}) seem to be new.
An example in Section~\ref{se:examples} illustrates  the dependence of the solutions to
the recalibration problem on assumptions of distribution shift and helps to identify some less reliable recalibration methods.
Section~\ref{se:conclusions} proposes a way forward to prudent recalibration and concludes the paper.
Appendix~\ref{se:App} provides additional technical details
needed for the implementation of some of the methods discussed in Section~\ref{se:approaches}.

%%%
\section{Literature review}
\label{se:relWork}
%%%

Calibration of probabilistic classifiers has often been treated in the literature, see the surveys by
\citet{calibratingOjeda2023} and \citet{silvaCalibration2023}. In the
typical calibration setting, a real-valued score  is learned on
a training dataset with joint observations of instances and labels. The score  is then \emph{calibrated} or 
\emph{mapped} to become a probabilistic classifier on a test (or validation or calibration) dataset 
for which there are also joint observations of instances and labels. If the original score is
already a probabilistic classifier then the term recalibration is sometimes used instead of calibration. 
Recent work by \citet{moreo2025interconnections} shows that, under distribution shift, the problem of classifier calibration is closely 
related to the problems of quantifying prior class probabilities and predicting the accuracy of a classifier.

Cautious calibration \citep{Allikivi2024cautious} is a variant of binary calibration
with the goal to avoid either overconfidence or underconfidence of probabilistic classifiers. Like for
calibration, it is assumed that the labels of the instances in the calibration and test datasets are known.
A common feature with recalibration as defined in this paper is conservatism of the posterior estimates. 
In the setting of this paper, conservatism can be achieved by choosing an appropriate value for
the target class-1 prior probability. 

Recalibration of probabilities of default (PDs) on a dataset without observation of labels but
with knowledge of the prior probability of positive labels has been a topic of research for twenty
or more years in credit risk \citep[][and the references therein]{Bohn&Stein}. 
Such recalibration may be considered
an extreme case of learning with label proportions \citep{QuadriantoLabelProportions}
where there are no individual label observations but label proportions for groups of instances are available. 
In general, `learning with label proportions' in the binary case 
requires that there are at least two groups of instances
with different proportions of positive labels such that results from the related research are not 
applicable to the recalibration problem as studied in this paper.

Quantification (or class distribution estimation, CDE) is another related problem. 
See \citet{esuli2023learning} for a recent survey. 
The goal of CDE in the binary case is to
estimate the proportion of positive labels in a test dataset without any information on the labels.
The primary common feature of the CDE and recalibration problems is the dependence of the solutions upon assumptions
on the type of distribution shift between the training and test datasets. 
However, Remark~\ref{rm:one} in Section~\ref{se:CSPD} shows that the one-parameter version of 
the recalibration method `parametric CSPD' may also be used for CDE.

Covariate shift with posterior drift (CSPD) as introduced by \citet{Scott2019} turns out
to be a useful assumption on the type of distribution shift for tackling the recalibration problem. 
See Sections~\ref{se:CSPD} and \ref{se:examples} below.

%%%
\section{Setting}
\label{se:setting}
%%%

In this paper, the focus is on the binary case, i.e.~two-class classification problems for which the following setting is assumed: 
\begin{itemize}
\item There are a class variable $Y$ with values in some measurable space $\mathcal{Y} = \{0, 1\}$ and  
a features (also called covariates) vector  $X$ with values in $\mathcal{X}$. 
Each example (or instance) to be classified has a class label $Y$ and features  $X$. 
\item In the training dataset, for all examples their features $X$ and labels $Y$ are observed. $P$ denotes the
training joint distribution, also called source distribution of $(X, Y)$ 
from which the training dataset has been sampled.
\item In the test dataset, only the features $X$ of an example can immediately be observed. Its label $Y$
becomes known only with delay or not at all. 
$Q$ denotes the test joint distribution, also called target distribution of $(X, Y)$ from 
which the test dataset has been sampled.
\item For the sake of a more concise notation, write for short $p = P[Y=1]$ and $q = Q[Y=1]$ and 
assume $0 < p < 1$ and $0 < q < 1$.
\end{itemize}
In addition, the notation $E_P[Z] = \int Z\, d P$ and $E_Q[Z] = \int Z \,d Q$ is used
for integrable real-valued random variables $Z$. 

The setting described above is called \emph{dataset shift} \citep{storkey2009training} or 
\emph{distribution shift} \citep{pmlr-v80-lipton18a}  
if source and target distribution of $(X,Y)$ are not the same, i.e.~in case of $P(X,Y) \neq Q(X,Y)$. 

\subsection{Recalibration} 
\label{se:recal}

Assume that since the joint source distribution $P(X,Y)$ of features 
$X\in \mathcal{X}$ and class $Y\in\{0,1\} = \mathcal{Y}$ 
is given, also the posterior probability 
$\eta_P(X) = P[Y=1\,|\,X]$ is known or can be estimated from the training dataset. $\eta_P(X)$ is a special case of 
probabilistic classifiers $\eta$ which are real-valued statistics with $0 \le \eta \le 1$ and
typically intended to approximate $\eta_P(X)$. Probabilistic classifiers for their part are
special cases of scores (or scoring classifiers) which are real-valued statistics intended 
to provide a ranking of the instances in the dataset in the sense that a high score suggests 
a high likelihood that the instance has class label~$1$ (positive label).

By assumption only the target marginal
feature distribution $Q(X)$ is observed while the target joint distribution $Q(X,Y)$ is unknown.
Nonetheless, for the recalibration problem, it is assumed that the target marginal label distribution, specified by 
$q = Q[Y=1]$ to be known. This is not in contradiction to $Q(X,Y)$ being unknown because in general
the ensemble of marginal distributions does not uniquely determine the joint distribution.
An example for this phenomenon is presented in Section~\ref{se:examples} below.
 
The goal is to fit a posterior probability $\eta_Q(x) = Q[Y=1\,|\,X=x]$ as
some transformation $T$ of $\eta_P(x)$ 
such that in particular it holds that
\begin{subequations}
\begin{equation}\label{eq:cond}
q \ = \ E_Q[\eta_Q(X)]\ = \ E_Q\bigl[T(\eta_P(X))\bigr].
\end{equation}
In the following, this problem is called \emph{recalibration} of the posterior 
probability $\eta_P(X)$ to a new
prior probability $q$ of class~$1$ under the target distribution.

Note that the assumption
\begin{equation}\label{eq:admissible}
\eta_Q(X) \ = \ T\bigl(\eta_P(X)\bigr)
\end{equation}
\end{subequations}
appears quite natural but actually is rather strong. Indeed,
by Theorems 32.5 and 32.6 of \citet{devroye1996probabilistic}, \eqref{eq:admissible}
is equivalent to $\eta_P(X)$ being sufficient for $X$ with respect to $Y$ under $Q$, i.e.
$Q[Y=1\,|\,X] \ =\ Q[Y=1 \,|\, \eta_P(X)]$.
This sufficiency property  may be interpreted as `the information provided by $\eta_P(X)$ 
about $Y$ is as good as the information from the whole set of features $X$ under the target distribution $Q$'.

\subsection{Non-uniqueness of recalibration}

The recalibration problem is not well-posed in the sense that
its solution is not unique. Therefore, a context is assumed where underestimating
$E_Q\bigl[C(\eta_Q(X))\bigr]$
for some fixed concave function $C: [0,1] \to \mathbb{R}$ ought to be avoided. In general, it holds that
(by Jensen's inequality and Lemma~1.2 of \citealp{ConcenLalley2013})
\begin{equation}\label{eq:bounds}
(1-q)\,C(0) + q\,C(1) \ \le \ E_Q\bigl[C(Z)\bigr] \ \le \ C(q)
\end{equation}
for any random variable $0 \le Z\le 1$ with $E_Q[Z] = q$, and $Z = \eta_Q(X)$ in particular. 
The maximum value $C(q)$ is taken for constant $Z = q$, the minimum value $(1-q)\,C(0) + q\,C(1)$ is
realised for $Z$ with $Q[Z=1] = q = 1 - Q[Z=0]$.

However, assuming a distribution shift between source $P$ and target $Q$ which results in a constant 
posterior probability $\eta_Q(x) = q$ appears too restrictive in most real world environments. In 
Section~\ref{se:examples} below, it is demonstrated that
assuming preservation of classification performance as measured by AUC (Area Under the Curve, see
next section) between 
source $P$ and target $Q$ strikes a sensible note between too restrictive and too tolerant assumptions for the
estimation of $E_Q\bigl[C(\eta_Q(X))\bigr]$ under the target features distribution.

%%%
\subsection{Area Under the Curve (AUC)}
\label{se:defAUC}
%%%

AUC (Area Under the
Curve\footnote{`Curve' refers to \emph{ROC} (Receiver Operating Characteristic).
See \citet{Fawcett2006ROC} and the references therein for the most common 
definitions of ROC and AUC.})
is a popular measure of performance of a binary classifier, i.e.\ AUC 
is considered an appropriate measure of the classifier's ability to predict
the true class label of an instance. See \citet{Chen2018Calibration} for related comments.
Since AUC plays an important role in some of the recalibration methods discussed in the following sections, 
in this section population-level 
(in contrast to sample-based) representations of AUC are presented for use in the following.

Let $S = h(X)$ be a score which is a function of the features with values in an ordered set $\boldsymbol{\mathcal{S}}$.
Define the class-conditional distributions $P_y$, $y \in \mathcal{Y} =
\{0,1\}$, of $S$ by 
$P_y[S \in M]  =  P[S \in M\,|\,Y=y]$, for all measurable $M \subset \boldsymbol{\mathcal{S}}$.

If the score $S$ is assumed to be large for instances with high likelihood to
have class $1$ and small for instances with high likelihood to
have class $0$, then the AUC for $S$ is defined as
\begin{subequations}
\begin{equation}\label{eq:MannWhitney}
AUC_S\ =\ P^\ast[S_1 > S_0] + \frac{1}{2}\,P^\ast[S_1 = S_0],
\end{equation}
where $P^\ast = P_1 \otimes P_0$ denotes the product measure of $P_1$ and $P_0$, and
$S_1$ and $S_0$ are the coordinate projections of the space 
$\boldsymbol{\mathcal{S}} \times \boldsymbol{\mathcal{S}}$ on which
$P^\ast$ is defined. As a consequence, $S_1$ and $S_0$ are independent and 
$P^\ast[S_y \le s] = P_y[S \le s]$ for $y \in \{0, 1\}$ and all 
$s \in \boldsymbol{\mathcal{S}}$. 

By Definition~\eqref{eq:MannWhitney}, on the one hand $AUC_S$ is 
``equivalent to the probability that
the classifier will rank a randomly chosen positive instance
higher than a randomly chosen negative instance'' \citep[][pp.~868]{Fawcett2006ROC}
if both of the 
class-conditional distribution functions of the score $S$ are continuous. On
the other hand, it holds that $AUC_S = \frac{1}{2}$ for any uninformative 
score $S$ -- i.e.~in the case $P_0 = P_1$ -- even if both of the class-conditional
score distribution functions have discontinuities. Note that computing $AUC_S$ by means
of \eqref{eq:MannWhitney} at first glance requires knowledge of the joint distribution $P(S,Y)$ of $S$ and $Y$
which would have to be inferred from a sample of paired $(S,Y)$ observations.

However, $AUC_S$ can also be determined if the distribution $P(S)$ of $S$ and the
posterior probabilities $\eta_P(S) = P[Y=1\,|\,S]$ are known. Then with $p  = E_P[\eta_P(S)]$ it holds that\footnote{%
The \emph{indicator function} $\mathbf{1}_A$ is defined by $\mathbf{1}_A(a) =1$ if
$a \in A$ and $\mathbf{1}_A(a) =0$ if $a \notin A$.\label{fn:ind}}
\begin{equation}\label{eq:Bayes}
\begin{split}
P_1[S \in M] &  \ = \ \frac{E_P\bigl[\eta_P(S)\,\mathbf{1}_M(S)\bigr]}{p} \qquad \text{and}\\
P_0[S \in M] &  \ = \ \frac{E_P\bigl[(1-\eta_P(S))\,\mathbf{1}_M(S)\bigr]}{1-p}
\end{split}
\end{equation}
\end{subequations}
for all measurable sets $M \subset \boldsymbol{\mathcal{S}}$. 
Plug $P_0$ and $P_1$ from \eqref{eq:Bayes} in \eqref{eq:MannWhitney} 
to compute $AUC_S$.

To indicate the way $AUC_S$ is computed, in this paper, the notion of AUC is used if $AUC_S$ is assumed to be computed 
or inferred by means of \eqref{eq:MannWhitney} irrespectively of the origin of $P_0$ and $P_1$. 
A reference to \emph{implied} AUC is used if $AUC_S$ is assumed to be computed by
means of \eqref{eq:Bayes} in combination with \eqref{eq:MannWhitney}. 
See Appendix~\ref{se:AUCauto} for a more detailed formula for implied 
AUC in the case of a discrete-valued score $S$.

%%%
\section{Approaches to recalibration}
\label{se:approaches}
%%%

Any solution as in \eqref{eq:admissible} to the recalibration problem 
together with the marginal feature distribution $Q(X)$ completely
determines the target distribution $Q(X,Y)$. Since $Q(X,Y) \neq P(X,Y)$ in case $q \neq p$, any given 
solution specifies some distribution shift between the source and target distributions. Hence, in order to
better understand the consequences of selecting a particular solution transformation $T$, it is natural
to explore the solution approaches through the lens of distribution shift. 
In the following, the assumption $0  < \eta_P(x) < 1$ for all $x \in \mathcal{X}$ is made.

\subsection{Recalibration under assumption of stretched or compressed covariate shift}
\label{se:covShift}

At first glance recalibration of $\eta_P(X)$ to a fixed target prior probability $q$ might appear to
be straightforward: Just define 
$\eta_Q(x)  = \frac{q}{E_Q[\eta_P(X)]}\,\eta_P(x)$ for $x \in \mathcal{X}$, 
then $E_Q[\eta_Q(X)] = q$ immediately follows. This approach is called \emph{scaling}
(Section~3.1 of \citealp{PtakRecalibration}).

Unfortunately, in some cases there is a problem with this approach:  $1 < \eta_Q(x)$ may be incurred in
the case $q > E_Q[\eta_P(X)]$.  To avoid this issue, one can modify the approach to become
\begin{subequations}
\begin{equation}\label{eq:capped}
\eta_Q(x) = \min\bigl(t\,\eta_P(x), 1\bigr),
\end{equation}
with $t >0$ being determined by 
\begin{equation}\label{eq:m}
q \ = \ E_Q\bigl[\min\bigl(t\,\eta_P(x), 1\bigr)\bigr].
\end{equation}
\end{subequations}
\eqref{eq:capped} could be called \emph{capped scaling} of the source posterior probabilities. Obviously,
\eqref{eq:capped} implies that \eqref{eq:admissible} is satisfied with $T(\eta) = \min(t\,\eta, 1)$.
 
Recall that source distribution $P(X,Y)$ and target distribution $Q(X,Y)$ are related through 
\emph{covariate shift} (in the sense of \citealp{storkey2009training}) 
if it holds that
$P[Y=y\,|\,X]  =  Q[Y=y\,|\,X]$
for all $y \in \mathcal{Y}$ with probability~$1$ both under $P$ and $Q$.
Hence if $t > 1$ in \eqref{eq:capped}, one might call the implied distribution shift 
\emph{stretched covariate shift}. 

In case $t <1$ the induced distribution shift could be considered \emph{compressed covariate shift}.
In case $q > E_Q[\eta_P(X)]$, \eqref{eq:capped} and \eqref{eq:m} imply $t > 1$ and $\eta_Q(x) = 1$ for all $x$ 
with $t\, \eta_P(x) \ge 1$. This may have the consequence of an undesired increase of 
$AUC_{\eta_Q(X)}$ compared to 
$AUC_{\eta_P(X)}$ for the area under the curve AUC defined as in Section~\ref{se:defAUC} above. 

\subsection{Recalibration under assumption of label shift}
\label{se:labelShift}

Source distribution $P(X,Y)$ and target distribution $Q(X,Y)$ are related through 
\emph{label shift} \citep{pmlr-v80-lipton18a}, previously called \emph{prior 
probability shift} in the literature \citep{storkey2009training}, if
$P[X \in M\,|\,Y=y] = Q[X \in M\,|\,Y=y]$
for all measurable sets $M \subset \mathcal{X}$ and $y\in\mathcal{Y}$.

Under the assumption of label shift, the target feature distribution $Q(X)$ 
can be represented as 
\begin{equation}\label{eq:mixture}
Q[X\in M]\ =\ q\,P[X \in M\,|\,Y=1] + (1-q)\,P[X \in M\,|\,Y=0],
\end{equation}
for all measurable sets $M \subset \mathcal{X}$. If a prior probability
$q = Q[Y=1]$ is given, then according to the \emph{posterior correction formula}
\citep[Eq.~(2.4) of][]{saerens2002adjusting}, 
$\eta_Q$ is determined through (recall $p = P[Y=1]$)
\begin{equation}\label{eq:correct}
\eta_Q(x) \ =\ \frac{\frac{q}{p}\,\eta_P(x)}{\frac{q}{p}\,\eta_P(x) +
	\frac{1-q}{1-p}\,(1-\eta_P(x))},
\end{equation}
for all $x\in\mathcal{X}$ with probability $1$ under $Q$.
In the credit risk community, the use of \eqref{eq:correct} for recalibration is popular 
(Section ``Calibrating to PDs'' of \citealp{Bohn&Stein}, Section~3.1 of 
\citealp{PtakRecalibration}) because it avoids the problem of $\eta_Q(X)$ potentially
taking the value $1$ for large $\eta_P(X)$ which may be encountered with capped
scaling as in \eqref{eq:capped}. \citet[][Sections~6.2 and 6.3]{Cramer2003}  pointed out that in 
the context of logistic regression the related special case of \eqref{eq:correct} was known at least since
1979.
Note that \eqref{eq:correct} implies \eqref{eq:admissible}
with $T(\eta) = \frac{\frac{p}{q}\,\eta}{\frac{p}{q}\,\eta + \frac{1-p}{1-q}\,(1-\eta)}$ strictly increasing in
$\eta$.

Under the label shift assumption, the following observation is well-known. 
Define AUC (area under the curve) as in Section~\ref{se:defAUC}. 
\begin{proposition}\label{pr:AUC}
Define the score $S$ by $S = \eta_P(X)$ and the 
score $S^\ast$ by $S^\ast = \eta_Q(X)$. If $P$ and $Q$ are related through label shift then it follows that
$AUC_S  = AUC_{S^\ast}$.
\end{proposition}
Proposition~\ref{pr:AUC} is a consequence of
\eqref{eq:MannWhitney} in Section~\ref{se:defAUC} as well as \eqref{eq:admissible} and \eqref{eq:correct}. 
According to Proposition~\ref{pr:AUC}, recalibration under the assumption of label shift leaves the implied
performance under the target distribution -- sometimes called
\emph{discriminatory power} in the credit risk management literature (e.g.,~\citealp{Bohn&Stein}) -- 
unchanged when compared to the 
implied performance under the source distribution.  In particular, Proposition~\ref{pr:AUC} implies
a necessary criterion for distribution shift to be label shift. Accordingly, in case $AUC_S \neq AUC_{S^\ast}$ 
source distribution $P$ and target distribution $Q$ cannot be related through label shift.

\subsection{Recalibration under assumption of factorizable joint shift (FJS)}
\label{se:FJS}

According to \citet{he2022domain} and \citet{tasche2022factorizable},
the source distribution $P(X,Y)$ and the target distribution
$Q(X,Y)$ are related through \emph{factorizable joint shift} (FJS) if 
there are functions $g:\mathcal{X} \to [0, \infty)$ and $b:\mathcal{Y} \to [0, \infty)$
such that $(x,y) \mapsto g(x)\,b(y)$ for $x \in \mathcal{X}$ and $y \in \mathcal{Y}$ is 
a density of $Q(X,Y)$ with respect to $P(X,Y)$, i.e.\ it holds that
$Q[(X,Y)\in M]  =  E_P\bigl[\mathbf{1}_M(X,Y)\,g(X)\,b(Y) \bigr]$
for all measurable sets $M \subset \mathcal{X}\times\mathcal{Y}$. 
See footnote~\ref{fn:ind} 
for the definition of the indicator $\mathbf{1}_M$.

By Corollary~4 of \citet{tasche2022factorizable} and subject to mild technical conditions, under FJS
the joint target distribution $Q(X, Y)$ is fully specified by the target feature distribtuion $Q(X)$ and
the class~1 posterior probability
\begin{subequations}
\begin{equation}\label{eq:fact.correct}
\eta_Q(X) \ = \ \frac{\frac{q}{p}\,\eta_P(X)}{\frac{q}{p}\,\eta_P(X) +
	\frac{1}{\varrho}\,\frac{1-q}{1-p}\,(1-\eta_P(X))},
\end{equation}
where 
$0 < \frac{p}{(1-p)\,E_Q\left[\frac{\eta_P(X)}{1-\eta_P(X)}\right]} \le \varrho \le 
	\frac{p}{(1-p)}\,E_Q\left[\frac{1-\eta_P(X)}{\eta_P(X)}\right]$ 
and $\varrho$ is the unique solution to the equation
\begin{equation}\label{eq:fact.determ}
q \ = \ E_Q\left[\frac{\frac{q}{p}\,\eta_P(X)}{\frac{q}{p}\,\eta_P(X) +
	\frac{1}{\varrho}\,\frac{1-q}{1-p}\,(1-\eta_P(X))}\right].
\end{equation}
\end{subequations}
Note that recalibration of $\eta_P(X)$ to a target prior probability $q$ under the assumption of FJS works for
arbritrary target feature distributions $Q(X)$. This is in stark contrast to recalibration under 
the assumption of label shift which only works when assuming that $Q(X)$ is given by \eqref{eq:mixture}.
However, by Proposition~\ref{pr:AUC} recalibration under the label shift 
assumption entails $AUC_{\eta_Q(X)} = AUC_{\eta_P(X)}$. The example
in Section~\ref{se:examples} below shows that AUC preservation  is not in general true for 
recalibration under the FJS assumption.

\eqref{eq:fact.correct} implies \eqref{eq:admissible}
with $T(\eta) = T_\varrho(\eta) = 
\frac{\frac{p}{q}\,\eta}{\frac{p}{q}\,\eta + \frac{1}{\varrho}\,\frac{1-p}{1-q}\,(1-\eta)}$  
strictly increasing in $\eta$. If $\varrho$ happens to take the value $1$, 
at first glance  the context of label shift as in 
Section~\ref{se:labelShift} above reappears. However, as \eqref{eq:mixture} need not hold true 
under the FJS assumption, in such cases there
cannot be label shift. The type of shift modelled instead is called `invariant density ratio' shift 
\citep{tasche2017fisher}, defined as the special case of FJS with $\varrho=1$.

\subsection{Recalibration under assumption of covariate shift with posterior drift (CSPD)}
\label{se:CSPD}

In Sections~\ref{se:covShift}, \ref{se:labelShift} and \ref{se:FJS}, 
the transformation $T$ of \eqref{eq:admissible} was identified  only after   a recalibration method had been introduced
which was designed under the assumption of one of three types of distribution shift, namely slightly modified covariate
shift, label shift, and FJS. In contrast, in this section, 
$T$ is defined first and then used to characterise the type of distribution shift implied by the recalibration method.

According to \citet{Scott2019}, the source distribution $P(X,Y)$ and the target distribution
$Q(X,Y)$ are related through \emph{covariate shift with posterior drift} (CSPD) if there is 
a \emph{strictly increasing transformation} $T$ such that \eqref{eq:admissible} holds true. 

As mentioned in Section~\ref{se:recal}, \eqref{eq:admissible} implies that $\eta_P(X)$ is sufficient
for $X$ with respect to $Y$ under the target distribution $Q$. 
Since under CSPD the transformation $T$ is strictly increasing,
\eqref{eq:admissible} also implies that $\eta_Q(X)$ and $\eta_P(X)$ are strongly comonotonic 
\citep{Tasche2022}. As a consequence, Kendall's $\tau$ and Spearman's rank correlation both take
the maximum value $1$ when applied to $(\eta_Q(X), \eta_P(X))$. This suggests that CSPD is a strong assumption
which might be less often true than one would hope for. Nonetheless, comonotonicity of score and
posterior probability is a common assumption in the literature. \citet{Chen2018Calibration} call
this assumption the \emph{rationality assumption}.

If labels are available in the test dataset, under the CSPD assumption 
the transformation $T$ of \eqref{eq:admissible} can be approximately
determined by means of isotonic regression. However, the general assumption for this paper is that there are
no label observations for the instances in the test dataset. Therefore,  a moment matching
approach is applied instead (quasi moment matching, QMM).

To be able to do so, consider \emph{parametric CSPD} where the transformation $T$ of \eqref{eq:admissible} is
specified as follows through some strictly increasing and continuous distribution function $F$ on the real line and 
parameters $a, b \in \mathbb{R}$:
\begin{subequations}
\begin{equation}\label{eq:parametric}
T_{a,b}(u) \ = \ F\bigl(a\,F^{-1}(u)+b\bigr), \qquad \text{for}\ 0 < u < 1.
\end{equation}
This is not a radically new approach but rather a variation of what was called regression-based calibration
by \citet{calibratingOjeda2023}.
Here are two examples for natural choices of $F$ in \eqref{eq:parametric}: 
\begin{itemize}
\item Logistic distribution function (inverse logit): $F(x) = \frac{1}{1 + \exp(- x)}$, $x\in\mathbb{R}$.
The parametric CSPD approach with inverse logit is often called `Platt scaling' in
the literature.
However, as pointed out by \citet{calibratingOjeda2023}, `Platt scaling' sometimes also refers to
the specification of $T$ as
\begin{equation}\label{eq:origPlatt}
T_{a,b}(u) \ = \ \frac{1}{1 + \exp(-(a\,u+b))}, \qquad \text{for}\ 0 < u < 1.
\end{equation}
In the following, the term \emph{Platt scaling} refers to \eqref{eq:origPlatt} and \emph{logistic CSPD}
refers to \eqref{eq:parametric} with inverse logit.
\item Standard normal distribution  function (inverse probit): $F(x) = \Phi(x)$, $x\in\mathbb{R}$.
This choice of $F$ is referred to as \emph{normal CSPD}.
\end{itemize}
The idea for \emph{quasi-moment matching (QMM)} with parametric CSPD is to determine the parameters 
$a,b \in \mathbb{R}$ by solving the following equation system:
\begin{equation}\label{eq:match}
q  = E_Q\bigl[T_{a,b}(\eta_P(X))\bigr]\quad\text{and}\quad
AUC_{\eta_P(X)}  = AUC_{T_{a,b}(\eta_P(X))},
\end{equation}
\end{subequations}
with $T_{a,b}$ as in \eqref{eq:parametric} or \eqref{eq:origPlatt} and AUC 
defined  in Section~\ref{se:defAUC}. More precisely, in \eqref{eq:match},
$AUC_{\eta_P(X)}$ is computed with respect to $P(X,Y)$ while $AUC_{T_{a,b}(\eta_P(X))}$ is computed as implied AUC
with respect to $Q(X)$. See Section~\ref{se:defAUC} for the different ways to compute AUC.

In \eqref{eq:match}, AUC works like a second moment of the posterior probabilities. But in contrast to its effect on the second moment or
the variance, the prior probability of the positive class has no effect on AUC. This makes the assumption 
of AUC invariance between source and target distributions more plausible \citep{Tasche2009a}.

\begin{remark}\label{rm:one}
One-parameter CSPD as in \eqref{eq:parametric} with $b =0$ can be used for class distribution estimation 
(also called quantification), i.e.~to determine an unknown class prior probability $q = Q[Y=1]$ under the target
distribution. In this case, instead of solving \eqref{eq:match} for two parameters $a$ and $b$, the following
equation is solved for parameter $a$ only:
\begin{subequations}
\begin{equation}
AUC_{\eta_P(X)}\  =\ AUC_{T_{a,0}(\eta_P(X))}.
\end{equation}
Then the mean of the resulting posterior probability $\eta_Q(X) = T_{a,0}(\eta_P(X))$ 
under the target feature distribution $Q(X)$ is 
computed to obtain an estimate $\widehat{q}$ of $q$:
\begin{equation}
\widehat{q}\ =\ E_Q\bigl[T_{a,0}(\eta_P(X))\bigr].
\end{equation}
\end{subequations}
One-parameter CSPD may be interpreted as a modification of quantification under the assumption of covariate shift.
In contrast to assuming covariate shift when AUC can differ between source distribution and target distribution, 
with one-parameter CSPD by construction source AUC and target AUC are equal.
\end{remark}

\subsection{Quasi moment matching based on parametrised receiver operating characteristics}
\label{se:robLogit}

There is no guarantee that the QMM approach presented in Section~\ref{se:CSPD} is feasible in the sense
that there exists a solution to equation system \eqref{eq:match} or that the solution is unique. This reservation
motivates the following alternative approach where instead of beginning with representation \eqref{eq:parametric} of
the posterior probabilities, the starting point is a parametrised representation of the 
receiver operating characteristic (ROC) curve 
associated with the target distribution $Q(X,Y)$. 

\citet[][Section~5.2]{Tasche2009a}  modified an idea of \citet{VanDerBurgt}
by assuming the ROC curve of a real-valued score $S$ to be 
\begin{subequations}
\begin{equation}\label{eq:ROC}
ROC_S(u) \ = \ \Phi\bigl(c + \Phi^{-1}(u)\bigr), \qquad u \in (0,1),
\end{equation}
for some fixed parameter $c \in \mathbb{R}$, with $\Phi$ denoting the standard normal distribution function.
The ROC curve of \eqref{eq:ROC} emerges when the class-conditional score distributions in 
a binary classification problem are both univariate normal distributions with equal variances.
But ROC curves like in \eqref{eq:ROC} may also be incurred in circumstances where the class-conditional
distributions are not normal (Proposition~5.3 of \citealp{Tasche2009a}).

\citet{Tasche2009a} showed that \eqref{eq:ROC} implies the following representation
for the posterior probability given the score $S$ under the target distribution $Q$:
\begin{equation}\label{eq:ROCLogit}
Q[Y=1\,|\,S] \ = \ \frac{1}{1 + \frac{1-q}{q}\,\exp\bigl(c^2/2 - c\,\Phi^{-1}(F_0(S))\bigr)},
\end{equation}
where $F_0(s) = Q[S\le s\,|\,Y=0]$ stands for the class-conditional distribution function of $S$ given $Y=0$.

For $AUC_S$ as defined in Section~\ref{se:defAUC}, \eqref{eq:ROC} implies 
$AUC_S =  \Phi\bigl(\frac{c}{\sqrt{2}}\bigr)$.
Hence, if $AUC_S$ is known the parameter $c$ of \eqref{eq:ROCLogit} is determined by
\begin{equation}\label{eq:cc}
c \ = \ \sqrt{2}\,\Phi^{-1}(AUC_S).
\end{equation}
\end{subequations}
The score $S$ in \eqref{eq:ROCLogit} can be chosen as $\eta_P(X)$ or any probabilistic
classifier which approximates $\eta_P(X)$. The target prior probability $q$ is known by general assumption for
this paper. Similarly to Section~\ref{se:CSPD}, for QMM to work one has to make 
the assumption that a prudent choice of $AUC_S$ under the target distribution
(defined as implied AUC by \eqref{eq:Bayes} and \eqref{eq:MannWhitney} above) is informed 
by $AUC_S$ observed in the training dataset such that by \eqref{eq:cc} also
parameter $c$ is known. 

However, the distribution function $F_0$ of the score $S$ conditional on
$Y=0$ appearing in \eqref{eq:ROCLogit} is assumed not to be known under the target distribution since 
by general assumption for this paper, only the features but not the labels
can be observed in the test dataset. Making use of \eqref{eq:ROCLogit} therefore 
requires an iterative approach where in each step a
refined estimate of the negative class-conditional score distribution function $F_0$ is calculated. 

Denote
by $f$ an unconditional density of $S$ under $Q$ and by $f_0$ a density of $F_0$. Then the
posterior probability $Q[Y=1\,|\,S=s]$ can be represented as
(see, e.g., Section~2.4.2 of \citealp{tasche2017fisher})
\begin{subequations}
\begin{equation}\label{eq:rep}
Q[Y=1\,|\,S=s]\  =\   1 - \frac{(1-q)\,f_0(s)}{f(s)}
\end{equation}
for $s$ in the range $\mathcal{S}$ of $S$. 
By \eqref{eq:ROCLogit}, \eqref{eq:rep} implies
\begin{equation}\label{eq:fixedPoint}
f_0(s)\ = \ \frac{f(s)}{1-q}\,\left(1-\frac{1}
{1 + \frac{1-q}{q}\,\exp\bigl(c^2/2 - c\,\Phi^{-1}(F_0(s))\bigr)}\right),
\quad s \in \mathcal{S}.
\end{equation}
\end{subequations}
In the case of scores $S$ under a discrete or empirical distribution, 
as described in Appendices \ref{se:AUCauto} and \ref{se:empDist},
\eqref{eq:fixedPoint} can be treated as a fixed point equation for the probabilities $Q[S=s\,|\,Y=0] = f_0(s)$
and be solved by a straightforward fixed-point iteration with intial values $Q[S=s]$, $s \in \mathcal{S}$.
The numerical
example of Section~\ref{se:examples} below suggests that such an iterative approach converges as long
as the prior probability $q$ is small and, hence, $f$ and $f_0$ are close to each other.

A further issue when making \eqref{eq:ROCLogit} operational may occur when the class-conditional
distribution function $F_0$ or any function approximating it takes the value $1$. In particular, this will happen
if $F_0$ is approximated by an empirical distribution function. Then the term $\Phi^{-1}\bigl(F_0(s)\bigr)$ is 
ill-defined. See Appendix~\ref{se:empDist} for a workaround to deal with this issue.

Deploying a discrete distribution $F_0$ in \eqref{eq:ROCLogit} as for instance an empirical approximation, 
makes it unlikely  if not impossible to exactly match a pre-defined $AUC_S$ when using \eqref{eq:ROCLogit}.
The unavoidable deviation from the $AUC_S$ objective in this case can be controlled with the 
original QMM approach as proposed by \citet{Tasche2009a}. Define
\begin{subequations}
\begin{equation}\label{eq:robLogit}
T^\ast_{a,b}(S) \ = \ \frac{1}{1 + \exp\bigl(b + a\,\Phi^{-1}(F_0(S))\bigr)},
\end{equation}
with $F_0$ as in \eqref{eq:ROCLogit} and parameters $a,b \in \mathbb{R}$ to be determined by quasi-moment
matching (QMM) like in \eqref{eq:match}:
\begin{equation}\label{eq:QMMorig}
q  = E_Q\bigl[T^\ast_{a,b}(S)\bigr]\quad\text{and}\quad
AUC_{S}  = AUC_{T^\ast_{a,b}(S)},
\end{equation}
\end{subequations}
To distinguish the two recalibration approaches presented in this section, the approach based
on \eqref{eq:ROCLogit} and \eqref{eq:cc} is called \emph{ROC-based QMM}. The approach based on
\eqref{eq:robLogit} and \eqref{eq:QMMorig} is referred to as \emph{2-parameter QMM}.

%%%
\section{Illustration}
\label{se:examples}
%%%

The illustrative example presented in this section shows the impact of the recalibration methods and 
assumptions discussed in Section~\ref{se:approaches} on the following characteristics of the
target distribution: 
\begin{itemize}
\item The class~1 posterior probabilities.
\item The mean of the class~1 posterior probabilities (which ought to equal the class~1 prior probability).
\item The AUC implied by the class~1 posterior probabilities.
\item The mean of the square root of the class~1 posterior probabilities as an example involving
a concave function of the posterior probabilities.
\end{itemize}
For the example discrete source and target distributions of a feature (called score in the following)
with values in an ordered set with 17 elements were chosen. These
distributions may be interpreted as empirical distributions of samples with many ties or as the
genuine distributions of discrete scores or ratings. For instance, the major credit rating agencies
Standard \& Poor's, Moody's and Fitch use rating scales with 17 to 19 different grades.

The source distribution is fully specified as follows:
\begin{itemize}
\item[(1)]  The conditional feature distribution for class~0 is a binomial distribution with success probability 0.4, the
conditional feature distribution for class~1 is a binomial distribution with success probability 0.55. 
The number of trials for both binomial distributions is 16 such that the support of the distribution is 
the set $\{0, 1, \ldots, 16\}$.
\item[(2)] The class~1 prior probability is $p=0.01$.
\end{itemize}

The target distribution is incompletely specified as follows:
\begin{itemize}
\item[(1)] The unconditional feature distribution is a binomial distribution with number of trials 16 whose
success probability is a Vasicek-distributed random variable (Section~2.2 of \citealp{meyer2009estimation}) 
with mean 0.3 and correlation parameter 0.3.
\item[(2)] The target class~1 prior probability is $q=0.05$.
\end{itemize}

\newpage

\begin{figure}[H]
\begin{center}
\ifpdf
	\includegraphics[width=14cm]{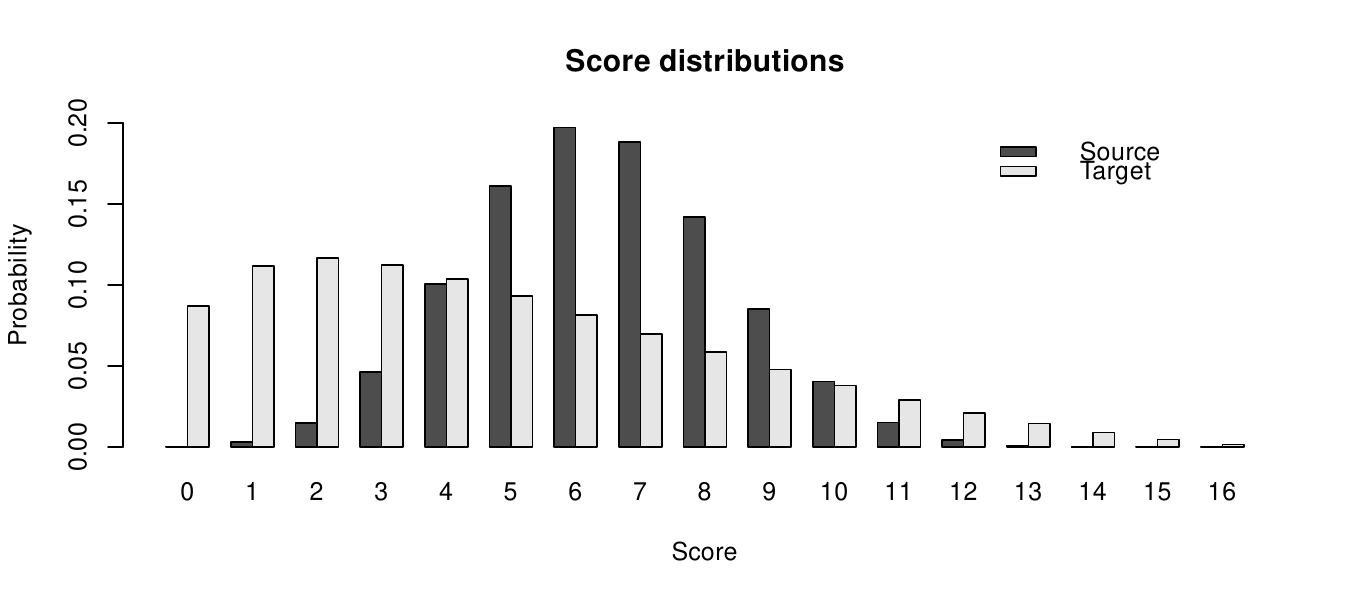}
%    \resizebox{\height}{14.0cm}{\includegraphics[width=20.0cm]{densities_1.pdf}}
\fi
\end{center}
\vspace{-2ex}
\caption{Source and target score distributions for the illustrative example.\label{fig:1}}
\end{figure}

Figure~\ref{fig:1} shows\footnote{%
Calculations were performed with R \citep{R2024}. Details
of some more involved calculations are described in Appendices~\ref{se:AUCauto} and \ref{se:empDist}.
The R-scripts can be downloaded from \href{https://www.researchgate.net/profile/Dirk-Tasche}%
{https://www.researchgate.net/profile/Dirk-Tasche}.
} the source and target unconditional score distributions. They
were intentionally chosen  to be quite different such that any distribution shift assumed for the recalibration
must be significant.

\begin{figure}[H]
\begin{center}
\ifpdf
	\includegraphics[width=14cm]{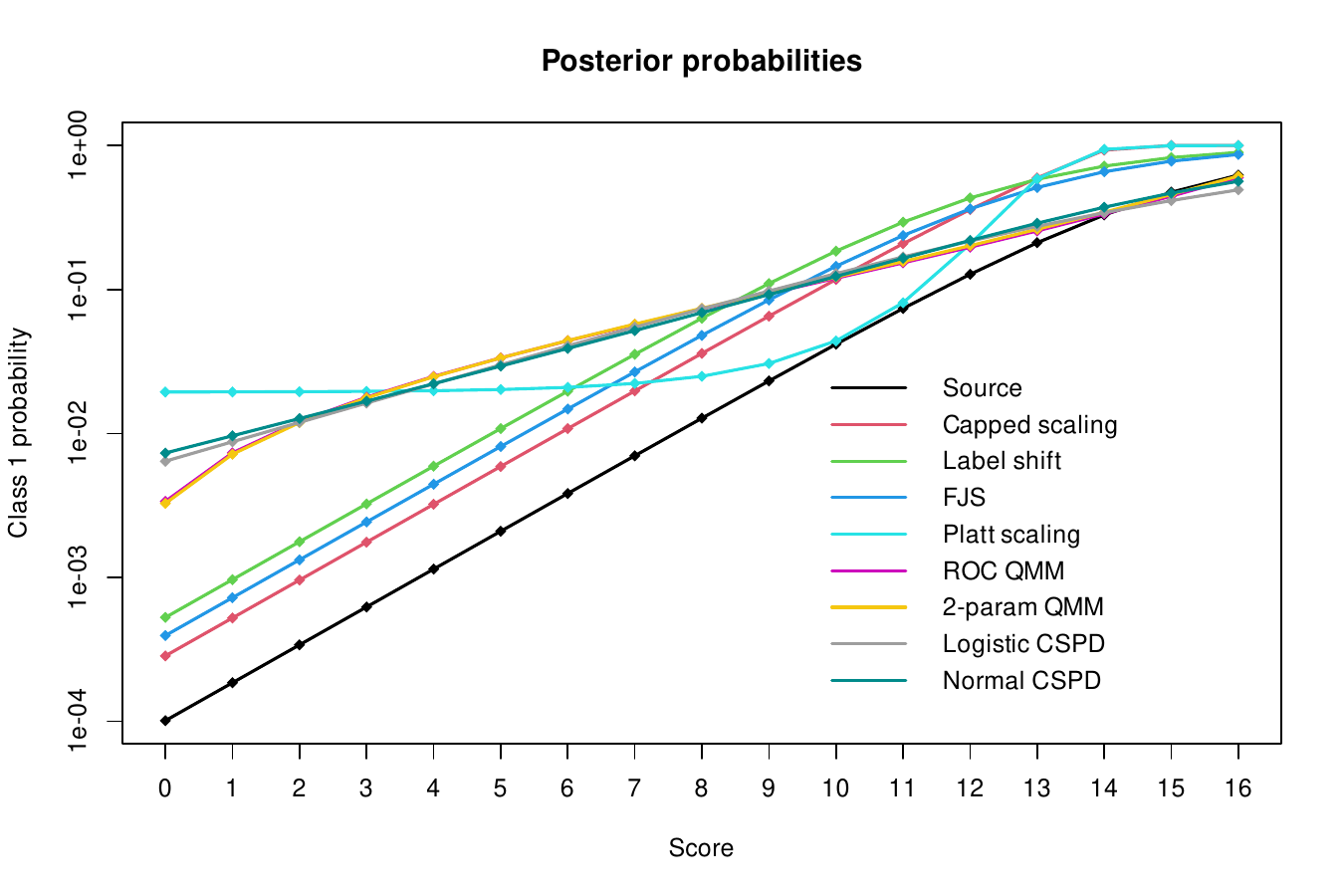}
%    \resizebox{\height}{14.0cm}{\includegraphics[width=20.0cm]{densities_1.pdf}}
\fi
\end{center}
\vspace{-2ex}
\caption{Class~1 posterior probabilities for the illustrative example.\label{fig:2}}
\end{figure}

Figure~\ref{fig:2} presents the source posterior probabilities and eight different sets
of target posterior probabilities that have been calculated with the methods and assumptions discussed in
Section~\ref{se:approaches}:
\begin{itemize}
\item Capped scaling: Section~\ref{se:covShift}
\item Label shift: Section~\ref{se:labelShift}
\item FJS: Factorizable joint shift, Section~\ref{se:FJS}
\item Platt scaling: Section~\ref{se:CSPD}, Eq.~\eqref{eq:origPlatt}
\item ROC QMM: ROC-based QMM, Section~\ref{se:robLogit}, Eq.~\eqref{eq:ROCLogit} and Eq.~\eqref{eq:cc}
\item 2-param QMM: 2-parameter QMM, Section~\ref{se:robLogit}, Eq.~\eqref{eq:robLogit} and Eq.~\eqref{eq:QMMorig}
\item Logistic CSPD: Section~\ref{se:CSPD}, Eq.~\eqref{eq:parametric} with $F(x) = \frac{1}{1 + \exp(- x)}$
\item Normal CSPD: Section~\ref{se:CSPD}, Eq.~\eqref{eq:parametric} with $F(x) = \Phi(x)$, the standard normal 
distribution function
\end{itemize}

In Figure~\ref{fig:2}, the
scale of the vertical axis is logarithmic. The dots for the probabilities have been connected with straight lines for better readability. 
`Source' refers to the posterior probabilities in the source distribution without recalibration.

All target posterior probabilities shown in Figure~\ref{fig:2} are well above the source posterior probabilities. This does not come as
a surprise given that the target class~1 prior probability of 5\% is much higher than the source class~1 
prior probability of 1\%.
Otherwise, there are two groups of target posterior probability curves: The curves based on capped scaling, label shift
and FJS on the one hand, and the curves based on the five QMM methods described in Sections~\ref{se:CSPD} 
and  \ref{se:robLogit} on the other hand. The curves of the former group are rather steep compared to the
curves of the latter group.

Table~\ref{tab:results} displays three characteristics for the source distribution as well as for the
eight different target distributions that result from the recalibration methods discussed in Section~\ref{se:approaches}.
The row `Source' shows the values for the source distribution without recalibration. 
The numbers in all other rows refer to the target distribution.
`mean(probs)' is the class~1 prior probability calculated as mean of the recalibrated
posterior probabilities. `AUC' is the area under the ROC curve implied by the recalibrated
posterior probabilities. `mean(sqrt(probs))' is the mean of the square root of 
the recalibrated posterior probabilities.

\begin{table}[H]
\begin{center}
\parbox{11cm}{\caption{Results for the application the recalibration methods to the illustrative example.}\label{tab:results}}
\begin{tabular}{|l||c|c|c|c|}\hline
\multicolumn{1}{|l||}{Method}&\multicolumn{1}{|c|}{\ mean(probs)}&\multicolumn{1}{|c|}{\quad AUC}&
\multicolumn{1}{|c|}{\ mean(sqrt(probs))}\\\hline\hline
Source&0.010&0.802&0.084\\\hline
Capped scaling&0.050&0.950&0.132\\\hline
Label shift&0.060&0.930&0.160\\\hline
FJS&0.050&0.932&0.142\\\hline
Platt scaling&0.050&0.802&0.179\\\hline
ROC QMM&0.049&0.799&0.191\\\hline
2-param QMM&0.050&0.802&0.191\\\hline
Logistic CSPD&0.050&0.803&0.192\\\hline
Normal CSPD&0.050&0.802&0.192\\
\hline
\end{tabular}
\end{center} 
\end{table}

Concluding from the entries in the column `mean(probs)' of Table~\ref{tab:results}, only the recalibration method `label shift' -- 
introduced in Section~\ref{se:labelShift} -- is unreliable in so far as it does not achieve the required 
target class~1 prior probability $p = 0.05$. With all seven other recalibration methods the target prior 
probability is matched or nearly matched. 

Column `AUC' of Table~\ref{tab:results} is more varied than column `mean(probs)'. For the five 
QMM methods from Sections~\ref{se:CSPD} and  \ref{se:robLogit}, AUC is essentially the same as the AUC of 0.802 under
the source distribution. This is a consequence of the QMM design since one of the moment matching objectives is hitting
the source AUC. Thanks to Proposition~\ref{pr:AUC}, one might expect also the `label shift' AUC to match the
source AUC. However, Proposition~\ref{pr:AUC} is not applicable to the example of this section because by design
the target feature distribution is not a mixture of the source class-conditional distributions. In any case,
the high AUC values for methods `capped scaling', `label shift' and `FJS' cause the greater slopes
of their posterior probability curves in Figure~\ref{fig:2} compared to the slopes of the
QMM posterior probability curves.

As an example for the application of a concave function $C$ to the target posterior probabilities and the related expected value
$E_Q\bigl[C(\eta_Q(X))\bigr]$ under the target distribution,
choose the function $C(u) = \sqrt{u}$, $u \in [0,1]$. Note that \eqref{eq:bounds} implies 
$0.05  =  q  \le  E_Q\bigl[\sqrt{\eta_Q(X)}\bigr] \le \sqrt{q}  \approx  0.2236$.

It is clear from column `mean(sqrt(probs))' of Table~\ref{tab:results} that 
AUC is a driver of the mean of the concave function of the posterior probabilities.
The lower the value of AUC, the higher is the mean of the concave function of the probabilities. Hence it makes sense to
have AUC as a second objective to be matched, in addition to requiring that the target class~1 prior
probability is reached. But even if both of these conditions are met there can still be variation
in the mean of the concave function. This is demonstrated by the `Platt scaling' posterior 
probabilities for which $E_Q\bigl[\sqrt{\eta_Q(X)}\bigr]$ takes a notedly lower value than for the other
four QMM posterior probabilities with almost identical values. 

%%%
\section{Conclusions}
\label{se:conclusions}
%%%

Recalibration of binary probabilistic classifiers to a target prior probability is an important task in areas like
credit risk management. This paper presents analyses and methods for recalibration from a distribution shift
perspective. In order to deal with the fact that there is no unique solution for the recalibration problem,  
the impact of the recalibration method on the values of concave or nearly concave functions 
of the posterior probabilities was investigated as an additional criterion
to identify meaningful solutions. It turned out that distribution shift assumptions linked to the performance of the 
probabilistic classifier in terms of the
area under the curve (AUC) under the source distribution are useful for
the design of useful recalibration methods. Two new methods called parametric CSPD (covariate shift
with posterior drift) and ROC-based QMM (quasi moment matching) were proposed and were tested 
together with some other methods in an example setting. The outcomes of this testing exercise suggest 
that the QMM methods
discussed in the paper can provide appropriately conservative results in evaluations with concave functions
like for instance risk weights functions for credit risk. 

\section*{Acknowledgement}

The author would like to thank the editorial team and an anonymous reviewer for their helpful feedback 
which greatly improved the quality of this paper.

%\bibliographystyle{apa}
%\bibliography{C:/Users/Dirk/Documents/LehreForschung/Papers/Literature}
%\bibliography{/home/dirk/Transfer2HP/Literature}

\begin{thebibliography}{}

\bibitem[\protect\astroncite{Allikivi et~al.}{2024}]{Allikivi2024cautious}
Allikivi, M.-L., J{\"{a}}rve, J., and Kull, M. (2024).
\newblock {Cautious Calibration in Binary Classification}.
\newblock {\em Frontiers in Artificial Intelligence and
  Applications}, 392, 1503--1510.

\bibitem[\protect\astroncite{BCBS}{}]{BCBSFramework}
BCBS.
\newblock {\em {CRE -- Calculation of RWA for credit risk}}.
\newblock Basel Committee on Banking Supervision,
\newblock {Regulatory Standard}, 
\newblock \url{https://www.bis.org/basel_framework/standard/CRE.htm}.

\bibitem[\protect\astroncite{Bohn and Stein}{2009}]{Bohn&Stein}
Bohn, J. and Stein, R. (2009).
\newblock {\em Active {C}redit {P}ortfolio {M}anagement in {P}ractice}.
\newblock John Wiley \& Sons, Inc.

\bibitem[\protect\astroncite{Chen et~al.}{2018}]{Chen2018Calibration}
Chen, W., Sahiner, B., Samuelson, F., Pezeshk, A., and Petrick, N. (2018).
\newblock Calibration of medical diagnostic classifier scores to the
  probability of disease.
\newblock {\em Statistical methods in medical research}, 27(5), 1394--1409.

\bibitem[\protect\astroncite{Cramer}{2003}]{Cramer2003}
Cramer, J. (2003).
\newblock {\em Logit {M}odels {F}rom {E}conomics and {O}ther {F}ields}.
\newblock Cambridge University Press.

\bibitem[\protect\astroncite{Devroye et~al.}{1996}]{devroye1996probabilistic}
Devroye, L., Gy{\"o}rfi, L., and Lugosi, G. (1996).
\newblock {\em A {P}robabilistic {T}heory of {P}attern {R}ecognition}.
\newblock Springer.

\bibitem[\protect\astroncite{Esuli et~al.}{2023}]{esuli2023learning}
Esuli, A., Fabris, A., Moreo, A., and Sebastiani, F. (2023).
\newblock {\em {Learning to Quantify}}.
\newblock Springer Cham.

\bibitem[\protect\astroncite{Fawcett}{2006}]{Fawcett2006ROC}
Fawcett, T. (2006).
\newblock An introduction to {ROC} analysis.
\newblock {\em Pattern Recognition Letters}, 27(8), 861--874.

\bibitem[\protect\astroncite{He et~al.}{2021}]{he2022domain}
He, H., Yang, Y., and Wang, H. (2021).
\newblock {Domain Adaptation with Factorizable Joint Shift}.
\newblock Presented at the ICML 2021 Workshop on Uncertainty and Robustness in
  Deep Learning,
\newblock \url{https://doi.org/10.48550/ARXIV.2203.02902}.

\bibitem[\protect\astroncite{Lalley}{2013}]{ConcenLalley2013}
Lalley, S. (2013).
\newblock Concentration inequalities.
\newblock Lecture notes,
\newblock \url{https://galton.uchicago.edu/~lalley/Courses/386/index.html}.

\bibitem[\protect\astroncite{Lipton et~al.}{2018}]{pmlr-v80-lipton18a}
Lipton, Z., Wang, Y.-X., and Smola, A. (2018).
\newblock {Detecting and Correcting for Label Shift with Black Box Predictors}.
\newblock In Dy, J. and Krause, A., editors, {\em Proceedings of the 35th
  International Conference on Machine Learning}, {\em Proceedings
  of Machine Learning Research}, 80, 3122--3130. 

\bibitem[\protect\astroncite{Meyer}{2009}]{meyer2009estimation}
Meyer, C. (2009).
\newblock Estimation of intra-sector asset correlations.
\newblock {\em The Journal of Risk Model Validation}, 3(3), 47--79.

\bibitem[\protect\astroncite{Moreo}{2025}]{moreo2025interconnections}
Moreo, A. (2025).
\newblock {On the Interconnections of Calibration, Quantification, and
  Classifier Accuracy Prediction under Dataset Shift},
\newblock \url{https://doi.org/10.48550/arXiv.2505.11380}.

\bibitem[\protect\astroncite{Ojeda et~al.}{2023}]{calibratingOjeda2023}
Ojeda, F., Jansen, M., Thi\'ery, A., Blankenberg, S., Weimar, C., Schmid, M.,
  and Ziegler, A. (2023).
\newblock {Calibrating machine learning approaches for probability estimation:
  A comprehensive comparison}.
\newblock {\em Statistics in Medicine}, 42(29), 5451--5478.

\bibitem[\protect\astroncite{Ptak-Chmielewska and
  Kopciuszewski}{2022}]{PtakRecalibration}
Ptak-Chmielewska, A. and Kopciuszewski, P. (2022).
\newblock {New Definition of Default Recalibration of Credit Risk Models Using
  Bayesian Approach}.
\newblock {\em Risks}, 10(1).

\bibitem[\protect\astroncite{Quadrianto
  et~al.}{2009}]{QuadriantoLabelProportions}
Quadrianto, N., Smola, A., Caetano, T., and Le, Q. (2009).
\newblock {Estimating Labels from Label Proportions}.
\newblock {\em Journal of Machine Learning Research}, 10(82), 2349--2374.

\bibitem[\protect\astroncite{{R Core Team}}{2024}]{R2024}
{R Core Team} (2024).
\newblock {\em R: A Language and Environment for Statistical Computing}.
\newblock R Foundation for Statistical Computing, Vienna, Austria.

\bibitem[\protect\astroncite{Saerens et~al.}{2002}]{saerens2002adjusting}
Saerens, M., Latinne, P., and Decaestecker, C. (2002).
\newblock Adjusting the {O}utputs of a {C}lassifier to {N}ew a {P}riori
  {P}robabilities: {A} {S}imple {P}rocedure.
\newblock {\em Neural Computation}, 14(1), 21--41.

\bibitem[\protect\astroncite{Scott}{2019}]{Scott2019}
Scott, C. (2019).
\newblock {A Generalized Neyman-Pearson Criterion for Optimal Domain
  Adaptation}.
\newblock In {\em Proceedings of Machine Learning Research, 30th International
  Conference on Algorithmic Learning Theory}, 98, 1--24.

\bibitem[\protect\astroncite{Silva~Filho et~al.}{2023}]{silvaCalibration2023}
Silva~Filho, T., Song, H., Perello-Nieto, M., Santos-Rodriguez, R., Kull, M.,
  and Flach, P. (2023).
\newblock Classifier calibration: a survey on how to assess and improve
  predicted class probabilities.
\newblock {\em Machine Learning}, 112(9), 3211--3260.

\bibitem[\protect\astroncite{Storkey}{2009}]{storkey2009training}
Storkey, A. (2009).
\newblock When {T}raining and {T}est {S}ets {A}re {D}ifferent: {C}haracterizing
  {L}earning {T}ransfer.
\newblock In Qui{\~n}onero-Candela, J., Sugiyama, M., Schwaighofer, A., and
  Lawrence, N., editors, {\em Dataset {S}hift in {M}achine {L}earning},
  chapter~1, 3--28. The MIT Press, Cambridge, Massachusetts.

\bibitem[\protect\astroncite{Tasche}{2009}]{Tasche2009a}
Tasche, D. (2009).
\newblock Estimating discriminatory power and {PD} curves when the number of
  defaults is small.
\newblock Working paper,
\newblock \url{https://doi.org/10.48550/arXiv.0905.3928}.

\bibitem[\protect\astroncite{Tasche}{2017}]{tasche2017fisher}
Tasche, D. (2017).
\newblock {Fisher Consistency for Prior Probability Shift}.
\newblock {\em Journal of Machine Learning Research}, 18(95), 1--32.

\bibitem[\protect\astroncite{Tasche}{2021}]{Tasche2022}
Tasche, D. (2021).
\newblock Calibrating sufficiently.
\newblock {\em Statistics}, 55(6), 1356--1386.

\bibitem[\protect\astroncite{Tasche}{2022}]{tasche2022factorizable}
Tasche, D. (2022).
\newblock {Factorizable Joint Shift in Multinomial Classification}.
\newblock {\em Machine Learning and Knowledge Extraction}, 4(3), 779--802.

\bibitem[\protect\astroncite{Tsyplakov}{2013}]{TsyplakovEvaluation}
Tsyplakov, A. (2013).
\newblock {Evaluation of Probabilistic Forecasts: Proper Scoring Rules and
  Moments},
\newblock \url{https://mpra.ub.uni-muenchen.de/45186/}.

\bibitem[\protect\astroncite{Vaicenavicius
  et~al.}{2019}]{vaicenavicius2019evaluating}
Vaicenavicius, J., Widmann, D., Andersson, C., Lindsten, F., Roll, J., and
  Sch{\"o}n, T. (2019).
\newblock Evaluating model calibration in classification.
\newblock In {\em The 22nd International Conference on Artificial Intelligence
  and Statistics (AISTATS) 2019, Naha, Okinawa, Japan, Proceedings
  of Machine Learning Research}, 89, 3459--3467. 

\bibitem[\protect\astroncite{Van~der Burgt}{2008}]{VanDerBurgt}
Van~der Burgt, M. (2008).
\newblock Calibrating low-default portfolios, using the cumulative accuracy
  profile.
\newblock {\em Journal of Risk Model Validation}, 1(4), 17--33.

\end{thebibliography}

\addcontentsline{toc}{section}{References}

\appendix

\section{Appendix}
\label{se:App}

\subsection{The AUC of discrete auto-calibrated probabilitistic classifiers}
\label{se:AUCauto}

How to calculate implied AUC for an auto-calibrated\footnote{%
The probabilistic classifier $S$ is \emph{auto-calibrated} if for all $s$ in the range of $S$ it
holds that $\mu[Y=1\,|\,S=s] = s$ (\citealp{TsyplakovEvaluation}, Section~2.2).
} probabilistic classifier $S$ under a joint distribution $\mu(S,Y)$ 
of $S$ and $Y \in \mathcal{Y} = \{0, 1\}$?
Here it is assumed that the distribution of $S$ under distribution $\mu$ is discrete, i.e.\ it is is given
by pairs of score values $s_i$ and their probabilities $\mu[S=s_i] = \pi_i >0$, $i=1, \ldots, n$.
Then it follows from the AUC-definition and properties in Section~\ref{se:defAUC} that
\begin{equation}\label{eq:discreteAUC}
AUC_S \ = \ \frac{\sum_{i=2}^n \pi_i\,s_i \Big(\frac{1}{2}\,\pi_i(1-s_i) + \sum_{k=1}^{i-1} \pi_k\,(1-s_k) \Big)}
	{\Big(1 - \sum_{j=1}^n \pi_j\,s_j \Big) \sum_{j=1}^n \pi_j\,s_j}.
\end{equation}
 \eqref{eq:discreteAUC} is used to compute implied AUC both under the source distribution for $S=\eta_P(X)$ with $\mu = P$ 
and under the target distribution for $S = \eta_Q(X) = T\bigl(\eta_P(X)\bigr)$ with $\mu =Q$. Note that
$\eta_P(X)$ under $P$ and $\eta_Q(X)$ under $Q$ are both auto-calibrated probabilistic 
classifiers (Proposition~1 of \citealp{vaicenavicius2019evaluating}). 

\subsection{Adapting discrete distributions for QMM}
\label{se:empDist}

Consider for the real-valued score or probabilistic classifier $S$  with 
discrete distribution $\mu[S=s_i] = \pi_i$, $i=1, \ldots, n$, its distribution
function $G(s) = \mu[S \le s]$ for $s \in \mathbb{R}$.
If the $s_i$ are increasingly ordered with $s_1 < \ldots < s_n$, it follows that
$G(s_i)  =  \sum_{j=1}^i \pi_j$, for $i = 1, \ldots, n$.
In particular, it follows $G(s_n) = 1$ such that $\Phi^{-1}\bigl(G(s_n)\bigr) = \infty$ for
the standard normal distribution function $\Phi$ and for the logistic distribution function.

For the purpose of this paper, to avoid this problem the workaround proposed by
\citet{VanDerBurgt} is applied. He suggested replacing the distribution function $G$  with
the mean of itself and its left-continuous version, i.e.~with $G^\ast$ defined by
$G^\ast(s)  =  \frac{1}{2}\,\bigl(\mu[S \le s] + \mu[S < s]\bigr)$, for $s \in \mathbb{R}$.

This implies for the $s_i$ which represent the support of $\mu$ that
$G^\ast(s_i)  =  \left(\sum_{j=1}^i \pi_j\right) - \pi_i / 2$, for $i = 1, \ldots, n$.

\end{document}